\documentclass[10pt,conference]{IEEEtran}
\IEEEoverridecommandlockouts
\usepackage{cite}
\usepackage{amsmath,amssymb,amsfonts}
\mathchardef\mhyphen="2D 
\usepackage{algorithm, float}
\usepackage[]{algpseudocode}
\usepackage{graphicx}
\usepackage{textcomp}
\usepackage{marginnote,colortbl}
\usepackage{boldline}
\usepackage{multirow, makecell}
\newcommand{\name}{{\tt DynaMIX}}
\def\BibTeX{{\rm B\kern-.05em{\sc i\kern-.025em b}\kern-.08em
    T\kern-.1667em\lower.7ex\hbox{E}\kern-.125emX}}
\newcolumntype{L}[1]{>{\raggedright\arraybackslash}p{#1}}
\newcolumntype{C}[1]{>{\centering\arraybackslash}p{#1}}
\usepackage{calc}
\usepackage{url}

\usepackage[dvipsnames]{xcolor}
\usepackage{breakurl}
\usepackage{tikz}
\newcommand*\circled[1]{\tikz[baseline=(char.base)]{
            \node[shape=circle,draw,inner sep=0.5pt] (char) {#1};}}
\newcommand*\fullcircled[4]{\tikz[baseline=(char.base)]{
    \node[shape=circle, fill=#2, draw=#3, text=#4, inner sep=1pt] (char) {#1};}}
\newcommand*\fullsmallcircled[4]{\tikz[baseline=(char.base)]{
    \node[shape=circle, fill=#2, draw=#3, text=#4, inner sep=0.5pt] (char) {#1};}}
\definecolor{midnightblue}{rgb}{0.1, 0.1, 0.44}

\begin{document}

\title{DynaMIX: Resource Optimization for DNN-Based Real-Time Applications 
on a Multi-Tasking System}

\author{\IEEEauthorblockN{Minkyoung Cho and Kang G. Shin \\
}
\IEEEauthorblockA{
\textit{The University of Michigan, Ann Arbor, MI, U.S.A.}\\
\{minkycho, kgshin\}@umich.edu}
}

\maketitle

\begin{abstract}
As deep neural networks (DNNs) prove their importance and 
feasibility, more and more DNN-based apps, such as detection and 
classification of objects, have been developed and deployed 
on autonomous vehicles (AVs).
To meet their growing expectations and requirements, AVs should 
``optimize" use of their limited
onboard computing resources for multiple concurrent in-vehicle apps 
while satisfying their timing requirements (especially for safety).
That is, real-time AV apps should share the limited on-board 
resources with other concurrent apps without missing their 
deadlines dictated by the frame rate of a camera 
that generates and provides input images to the apps. 
However, most, if not all, of existing DNN solutions focus on enhancing the 
concurrency of their specific hardware without dynamically 
optimizing/modifying the DNN apps' resource requirements, 
subject to the number of running apps, owing to their high computational cost.
To mitigate this limitation, we propose {\bf \name} 
(\underline{\tt Dyna}mic
\underline{\tt MIX}ed-precision model construction), which optimizes the resource requirement 
of concurrent apps and aims to maximize execution accuracy. 
To realize a real-time resource optimization, we formulate an 
optimization problem using app performance profiles to consider 
both the accuracy and worst-case latency of each app.
We also propose dynamic model reconfiguration by 
lazy loading only the selected layers at runtime to reduce 
the overhead of loading the entire model.
\name\ is evaluated in terms of constraint satisfaction and 
inference accuracy for a multi-tasking system and 
compared against state-of-the-art solutions, demonstrating 
its effectiveness and feasibility under various 
environmental/operating conditions.

\end{abstract}


\section{Introduction}
\begin{figure}[t]
    \begin{center}
    \vspace*{-3mm}
    \includegraphics[width=1\linewidth]{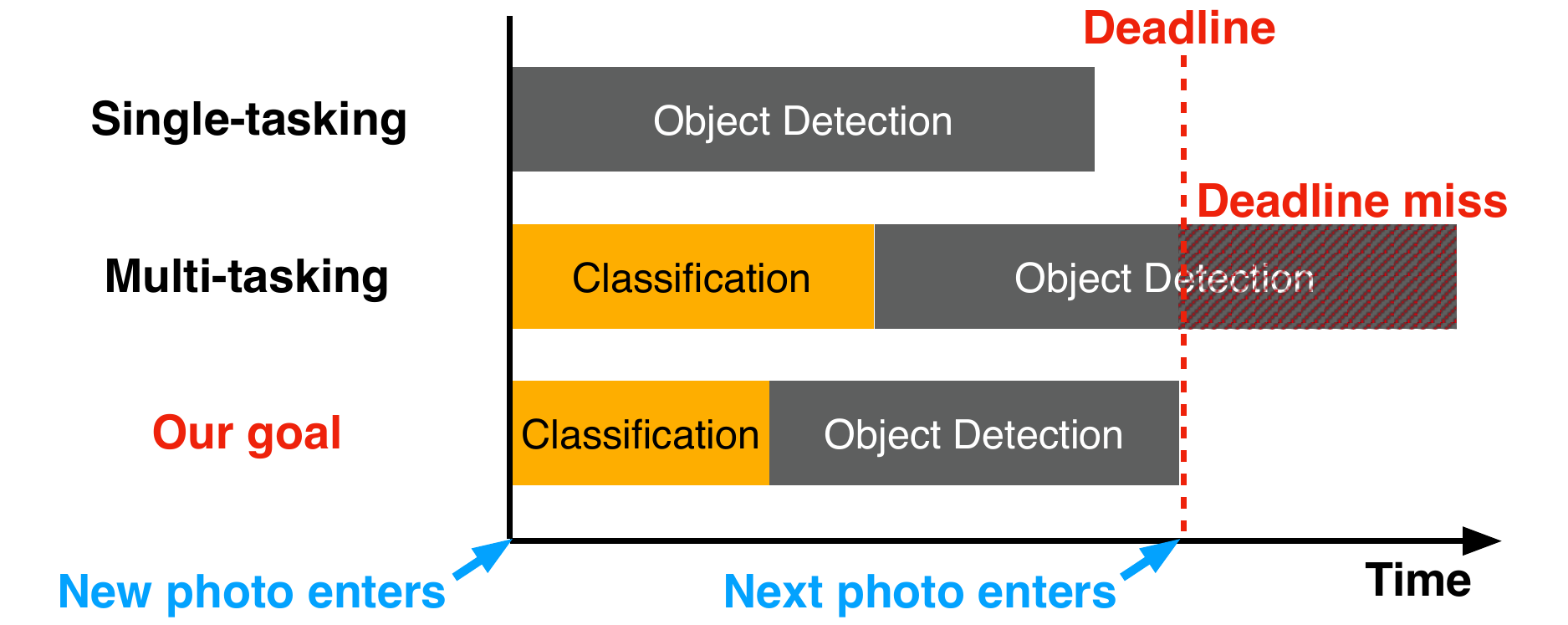} 
    \end{center}
    \caption{Example use-case and the objective of \name}
    \label{fig:objective}
    \vspace*{-3mm}
\end{figure}

As more deep neural network (DNN) apps are getting added in 
autonomous vehicles (AVs), it is critically important to 
ensure the timeliness and quality of their execution. 
Major car-makers, like Tesla \cite{tesla2021autopilot} and 
Toyota \cite{toyota}, are expanding the scope of using DNNs 
for camera-based vision apps.
In AVs, a camera continuously monitors environmental 
conditions at a rate of 10 -- 40 
frames per second (FPS) \cite{yang2019re, lin2018architectural}.
Since the results of processing the current image frame 
should be produced before the next frame arrives,
all in-vehicle apps using the image frames captured by a camera must 
meet the same deadline dictated by the camera's frame rate.
However, the interference between concurrent vision apps may lead to miss 
some of their deadlines.
Fig.~\ref{fig:objective} shows a motivating scenario in which two 
apps process image frames from the same camera. 
In such a case, the two apps have the {\em same} 
deadline since they must neither produce outdated data nor 
interfere with the processing of the next image frames.
If the execution of a classification app precedes that of 
a real-time object detection (RTOD) app, the end-to-end 
(i.e., camera-to-detection output) latency of the object 
detection app exceeds its deadline.

\begingroup
\setlength{\tabcolsep}{4pt}
\renewcommand{\arraystretch}{1}
\begin{table*}[tp]
    \small
    \caption{Comparison with prior work. ``Multi-tasking Support'' denotes running various apps not constrained by specific DNN models}
    \centering
    \vspace*{-2mm}
    \begin{center}
        \fontsize{9}{12}
        \begin{tabular}{c!{\vline width 1pt}ccc}
        \hlineB{2}
                                                                                               & Multi-Tasking Support & Real-Time Guarantee  & No Binding to Specific Processor \\ \hlineB{2}
        \multirow{2}{*}{\begin{tabular}[c]{@{}c@{}}A. Resource Management\\ \cite{han2016mcdnn, huynh2017deepmon, fang2018nestdnn} \end{tabular}} & \multirow{2}{*}{\checkmark}     & \multirow{2}{*}{}    & \multirow{2}{*}{\checkmark}           \\ 
                                                                                               &                       &                      &                             \\\hline
        \multirow{2}{*}{\begin{tabular}[c]{@{}c@{}}B. RTOD Execution\\ \cite{liu2020continuous, yang2019re, heo2020real, kang2022dnn} \end{tabular}}      & \multirow{2}{*}{}     & \multirow{2}{*}{\checkmark}    & \multirow{2}{*}{\checkmark}           \\
                                                                                               &                       &                      &                             \\\hline
        \begin{tabular}[c]{@{}c@{}}C. Hardware-based Scheduling\\ \cite{baek2020multi, jiang2020optimized, jeong2022band, han2022microsecond} \end{tabular}            & \multicolumn{1}{c}{\checkmark}  & \multicolumn{1}{c}{\checkmark} & \multicolumn{1}{c}{}        \\ \hline
        DynaMIX                                                                                &                \checkmark       &       \checkmark               &             \checkmark                \\\hlineB{2}
        \end{tabular} 
        \vspace*{-4mm}
    \end{center}
    \label{tbl:relatedwork}
\end{table*}

\looseness=-1
However, guaranteeing the deadline of embedded control apps 
like AVs remains a challenging problem.
When the number of concurrent apps changes dynamically, 
the amount of resources (e.g., CPU and memory space) 
they require should also change to meet their deadline.
Unfortunately, most vision apps are computation-intensive, 
and hence state-of-the-art (SOTA) multi-tasking systems
\cite{yang2019re, jiang2020optimized, jeong2022band, han2022microsecond}
focused on concurrent execution of multiple DNN apps
on specific processors without dynamically modifying
their resource requirements.
The most straightforward way to reduce the amount of resources 
required by each app is compressing the DNN model.
Given the tradeoff between accuracy and computation, 
uniformly compressing all the layers in a model 
with a single precision is not efficient.
For example, quantizing a model with high precision may result 
in a deadline miss. Conversely, low precision quantization 
will deteriorate accuracy \cite{krishnamoorthi2018quantizing}, 
which may cause fatal accidents.
This tradeoff calls for a design that optimizes each DNN model 
to be represented in mixed-precision while considering 
{\em total accuracy} (defined as the weighted sum of 
accuracies of concurrent apps) as well as its deadline.

\looseness=-1
In this paper, we propose \name\ (\underline{\tt Dyna}mic
\underline{\tt MIX}ed-precision model construction),
which dynamically changes the bit-width setting of each DNN model 
according to the status of running apps to optimize their 
computation and memory resources.
For the realization of \name, we must address two difficulties 
that make mixed-precision models unsuitable for 
time-critical systems like AVs.
First, we have to determine the optimal degree of compression 
(i.e., optimal bit configuration) for each DNN model at 
runtime. Balancing between latency and accuracy at runtime 
is difficult as both require to run DNN inferences 
as many as the number of compressed models of each app
when the status of running apps changes,
thus incurring high online overhead.
To address this difficulty,
we create measurements-based app performance profiles 
via regression for worst-case latency and most-likely 
inference accuracy according to the available memory size.
The resulting profiles are then used for online resource 
optimization across multiple apps.

Second, we need to reduce the overhead of model loading.
When their optimal bit settings are modified, 
loading the entire mixed-precision models is inefficient 
unless the bit-widths of all their layers are changed. 
However, loading {\em all} possible mixed-precision 
models generated from one DNN model into memory is neither 
efficient nor possible.
To reduce this model switching overhead,
we propose runtime model reconfiguration based on the concept of lazy loading,
commonly used in other domains such as web services and 
operating systems \cite{park2019design, arendt1998system}.
The proposed layer-wise lazy loading helps avoid the need 
for loading the entire model whenever the optimal bit setting 
is changed to use the platform resources efficiently.

\name\ is composed of offline and online stages.
In the online stage, it formulates the resource optimization 
as a constrained nonlinear optimization problem.
The constraints of the optimization include not only 
the deadline but also the offline-generated profiles to 
determine an optimal solution (i.e., memory space for 
each app) that maximizes the total accuracy subject to 
the deadline and the available memory space. 
When a new camera image enters the system and any change 
in the status of concurrent apps is detected, the main 
process adjusts the amount of memory space allocated 
to the concurrent apps.
After determining the new bit configuration with the 
adjusted memory space for each DNN app, the app 
reconfigures its execution path by loading the necessary 
high-precision layers to optimize the amount of 
computation resource. This way, all of them can meet 
the deadline without compromising the total accuracy.

We have evaluated \name\ for running multiple concurrent apps.
Our findings help one or more concurrent apps meet the 
deadline while optimizing the resource allocation. 
The resulting model for each app is shown 
to lose $<2.2$\% accuracy even in the case of an increased 
number of concurrent apps or tight timing constraints.
Furthermore, we have shown the feasibility of \name's 
resource allocation in various aspects and
demonstrated its effectiveness in handling multi-tasking 
scenarios. We have also identified the cases 
\name\ could not support, or the resulting total accuracy 
was too low to be acceptable for the apps in reality.

In summary, this paper makes the following contributions:

\begin{itemize}
    \item The first resource optimization scheme for 
       concurrent in-vehicle apps to meet their deadline 
       without binding to specific processors;
    \item Formulation of the optimization problem using app 
        performance profiles to find the optimal compression 
        degrees for multiple apps while accounting for 
        their accuracies and worst-case latencies in real time;
	\item Runtime model reconfiguration while lazy loading 
	  only the selected layers when the optimal bit setting 
	  of an app varies, to mitigate the entire model 
	  loading overhead; 
    \item Facilitation of rapid deployment of existing DNN 
      apps without any (re)training required to prevent any 
      significant accuracy drop in runtime adaptive 
      quantization.
\end{itemize}

\section{Related Work}
\label{sec:relatedwork}

Prior work on multiple or real-time DNN inferences on embedded 
platforms has evolved into three main branches: 
resource management, RTOD optimization, 
and hardware-based scheduling
(Table~\ref{tbl:relatedwork}).

\subsection{Resource Management for Multiple Vision Apps}

Resource management plays a critical role in ensuring 
the continuous execution of multiple vision apps.
Most resource-management schemes reduce the excessive cost of 
DNN pipeline execution by fully/partially offloading the 
computation to a cloud server or other edge devices. 
MCDNN \cite{han2016mcdnn}, a representative 
resource-management scheme, proposes a runtime scheduler 
to process DNN models (or fragments thereof) 
across both the edge device(s) and the cloud server. 
Although offloading reduces the computation 
load of the edge system, it yields inconsistent 
app quality because of unpredictable cloud accessibility 
or possible privacy leakage.

To mitigate/avoid these problems, \cite{huynh2017deepmon,fang2018nestdnn} 
suggested on-device resource management.
DeepMon \cite{huynh2017deepmon} aims to guarantee 
continuous vision apps by optimizing the convolutional neural 
networks (CNN) on mobile GPUs. It accelerated the convolution 
by reusing the intermediate results via caching.
NestDNN \cite{fang2018nestdnn} proposed a 
filter pruning for resource management.
However, both DeepMon and NestDNN
support only non-real-time tasks, probably because the 
most commonly used real-time tasks (e.g., object detection 
or tracking) require significant amounts of computation 
that the edge system cannot complete in a timely 
manner (e.g., DeepMon shows only 1$\sim$2 FPS).
\name\ was inspired by \cite{huynh2017deepmon, fang2018nestdnn} 
in that both pursue a server-less method.
\name\ reduces the complexity of vision apps via DNN 
compression, and finds the optimal models by exploiting 
the knowledge of app performance.

\subsubsection{Why not adapting NestDNN \cite{fang2018nestdnn} 
to guarantee deadlines?}
Even if NestDNN’s scheduler is revised to meet real-time 
constraints, it cannot guarantee both accuracy and timeliness.
NestDNN leverages a small, fixed number (i.e., five) 
of pruned models for each app, thus sacrificing accuracy 
to create models of various sizes.
If more pruned models were generated
for accuracy, timely execution is difficult because of its 
high overhead of finding an optimal model set for concurrent DNN apps.
To address this, they need to formulate an optimization 
problem using app performance profiles and generate 
lookup tables as we proposed here.
In contrast, \name\ fully supports real-time scenarios.

\subsection{RTOD Execution on Embedded Systems}
\label{sec:rtod}
The great potential of RTOD for improved safety and 
convenience has yielded a large body of work running 
RTOD on embedded systems, particularly AVs. Despite 
the significant recent advances in object detection 
\cite{liu2016ssd, ren2016faster, redmon2016you}, 
RTOD still remains to be a bottleneck in embedded 
real-time systems due to its heavy computation requirement
to localize and classify multiple objects 
captured in a stream of camera images.

\looseness=-1
AdaVP \cite{liu2020continuous} 
built a parallel object detection and tracking 
pipeline to run two apps in parallel; when a new object 
is detected, the DNN setting (i.e., input frame size) 
is adjusted at runtime to increase the tracking accuracy.
Yang {\em et al.} \cite{yang2019re} 
handled simultaneous vision apps with private 
camera streams and identical DNN models; throughput was 
increased by sharing the base architecture and 
processing multiple camera images with multiple threads.
DNN-SAM \cite{kang2022dnn} enabled multiple inferences 
with one RTOD model to process the images produced by 
different cameras. 
It ensures the deadline by splitting the original image 
into different portions according to the criticality levels 
and adjusts the size of each portion via scaling.

\looseness=-1
The authors of \cite{heo2020real} proposed a new RTOD 
system that could change the execution path based on 
a \textit{dynamic deadline} --- the deadline of a RTOD task 
varies with the underlying driving environment, like skipping 
layers or choosing one of the sub-networks of different sizes.
Moreover, they determined the execution path of RTOD 
using per-layer latency (for all layers) and 
dynamic deadline, but did not account for the 
interruptions by other concurrent apps in their 
path-selection decisions.
\name\ was inspired by \cite{heo2020real} 
in that the execution path can be changed dynamically
to account for the execution environmental condition.
\name\ reconfigures the DNN execution pipeline 
once the amount of available
resources for each app changes.

In summary, prior work accelerates the
RTOD task to meet its explicit/implicit deadline 
using model adaptation or multi-threading.
However, the resource contention by multiple threads/apps
may lead to performance degradation, and the real-time model 
adaptation can cause a significant accuracy drop.
More importantly, their requirement of full use 
of hardware resources for RTOD can be problematic when 
applied to a real-time multi-tasking platform.

\subsubsection{Why cannot Heo {\em et al.} \cite{heo2020real}~be 
modified for multi-neural network execution?}
\looseness=-1
The authors of \cite{heo2020real} showed that the WCET model 
for a certain layer can work in multi-tenant systems, 
but they neither showed how to share the limited resources 
between concurrent DNN models nor accounted for the 
overhead of context switching between apps. 
In contrast, \name\ addresses these problems in 
multi-tasking systems.

\subsection{Hardware-based Scheduling for Multi-DNN Inferences}
\looseness=-1
AI-MT \cite{baek2020multi} proposed a new NN accelerator 
architecture and scheduling scheme for multi-tasking platforms.
\cite{jiang2020optimized, jeong2022band} 
coordinated multiple latency-critical DNN tasks 
by using specific DNN accelerators (e.g., FPGA, NPU).
Scheduling jobs across heterogeneous processors 
allowed their real-time execution.
The most recent work in \cite{han2022microsecond} proposed 
a GPU scheduling method to run multiple real-time apps. 
Basically, it schedules memory- and computation-bound jobs 
across heterogeneous or several processors to enhance
concurrency. However, it does not work in the 
representative case of Fig.~1.
It only focuses on scheduling multiple real-time jobs with 
the unmodified DNN models which require a {\em static} 
amount of resources, thus limiting the number of 
concurrent jobs it can handle. 
In contrast, \name\ dynamically adjusts the resource 
consumption of each app. Moreover, \name\ is orthogonal
to these prior solutions, and hence can
run with them together.


\section{Background: Mixed-precision Quantization}
\label{sec:background}
\looseness=-1
Quantization of activation and model parameters can accelerate 
DNN execution by reducing the computational complexity of 
the underlying models. This is predicated on the fact 
that integer operations yield a much higher throughput in 
vectorized computations than floating-point operations 
on most computing platforms.
However, removing some bits in a fully-trained model 
(that is already converged to the lowest loss) causes an 
output perturbation between the full-precision and 
quantized models, thus degrading accuracy significantly.

\looseness=-1
Mixed-precision quantization has been explored as 
a promising solution to this problem by using layers of 
different bit-widths \cite{wang2019haq, cai2020zeroq, 
dong2019hawq, lee2021novel, zhe2019optimizing}. 
Using a higher bit-width at a layer more sensitive to 
quantization can help the layer preserve its original 
values, thus making the model suffer less output 
perturbation and accuracy drop.
Typical mixed-precision quantization algorithms are 
composed of layer sensitivity measurement and layer 
bit-width decision-making.

\subsection{Measurement of Layer Sensitivity}
\label{sec:sensitivity}

{\em Layer sensitivity} represents the extent to which the 
model output changes when a certain layer is quantized.
For the most exhaustive mixed-precision approach, a model 
of $L$ layers with $B$ types of bit-widths yields 
$B^L$-quantized models. Since DNN models are recently 
becoming deeper, mixed-precision will be less attractive.
So, identifying layer sensitivity is an efficient way to 
reduce the large design space for bit allocation.
Well-measured layer sensitivity can also be used to 
calculate the overall perturbation of the resulting 
mixed-precision model, and is a good measure for finding 
the best bit setting.

Layer sensitivity is affected by several factors, such as 
layer position, operation type, connection with other 
layers, and layer parameter size.
It is difficult to define sensitivity by considering 
all these factors in a large model; researchers 
used output perturbation as their sensitivity metric 
by calculating \textit{L2-Norm} or \textit{KL-Divergence}
\cite{zhe2019optimizing, cai2020zeroq}.

\looseness=-1
In this paper, we use the sensitivity metric 
defined in the state-of-the-art mixed-precision quantization 
method ZeroQ \cite{cai2020zeroq}, which is based on 
\textit{KL-Divergence} between the full-precision  
model and the quantized model, as:
\begin{align*}
  {S}_{i}(k)=\frac{1}{N} \sum_{j=1}^{N} \mathrm{KL}
  \big(\mathcal{N} \big(x_{j}\big), ~ \mathcal{\widetilde{N}}_{i}^{k}\big(x_{j}\big)\big)
\end{align*}
where ${S}_{i}(k)$ denotes the sensitivity of quantized 
model $\mathcal{\widetilde{N}}_{i}^{k}(\cdot)$ in which 
the $i$-th layer is quantized into $k$-bits, 
$\mathcal{N}(\cdot)$ indicates the full-precision model, 
and $x$ is a small set of input images of size $N$ used 
for sensitivity measurement.

\subsection{Hardware-aware Bit-width Decision}
\looseness=-1
Mixed-precision quantization has the flexibility of hardware-aware 
model compression. As the most basic method of bit-width decision, 
rule-based schemes have been used based on the knowledge of the 
DNN model and hardware architecture as well as manual effort.
However, increasingly complex models
make such heuristics difficult to apply.
ZeroQ \cite{cai2020zeroq} employed a 
\textit{Pareto frontier}-based method that finds an optimal 
compressed model with minimum output perturbation.
The overall sensitivity for each mixed-precision model is 
computed by summing the sensitivities of all layers in the model.
Although the authors did not reflect inter-layer dependency 
in this process, they showed that such sensitivity calculation 
incurs less computational overhead and produces good 
empirical results.
Additionally, sensitivity and model size are considered in 
ZeroQ as the indirect indicators of accuracy 
and latency. 

However, each processor has a distinct architectural design, 
implying that the best bit settings vary with hardware 
\cite{wang2019haq}.
For instance, a weight parameter layout designed to increase 
reusability can efficiently reduce the latency of conventional 
convolution layer, which can thus have a higher bit-width than 
fully-connected or depth-wise convolution layers.
Thus, with hardware-software codesign, this advanced quantization 
can reduce the computation and resource costs without 
any severe loss of inference accuracy compared to 
hardware-oblivious DNN compression methods (e.g., 
uniform quantization \cite{krishnamoorthi2018quantizing}).

Thanks to its flexibility, mixed-precision can accelerate the 
network model with a negligible accuracy degradation. 
Thus, recent processing engines have been 
released with the function of mixed-precision arithmetic 
with variable bit-widths \cite{nvidia_h100, raha20208}. 
These recent advancements have raised the importance of 
mixed-precision-based methods for DNN acceleration.

Inspired by the reinforcement learning-based bit-width setting 
method \cite{wang2019haq}, we determine the best mixed-precision 
model for each app in a hardware-aware manner.
Instead of using indirect indicators, e.g., FLOPs, model size, and sensitivity,
we use the worst-case latency (to meet 
the deadline), peak memory usage (to consider memory 
capacity), and inference accuracy measured in 
our simulated environment.
We aim to use a mixed-precision approach to find an optimal 
solution that maximizes the total accuracy while making the 
most of limited resources.
Sec.~\ref{sec:runtime-change} and~\ref{sec:architecture} will 
detail how \name\ builds the mixed-precision 
model and chooses an  optimal bit configuration 
for each DNN-based app.

\begin{figure}[tp]
    \begin{center}
    \includegraphics[width=0.9\linewidth]{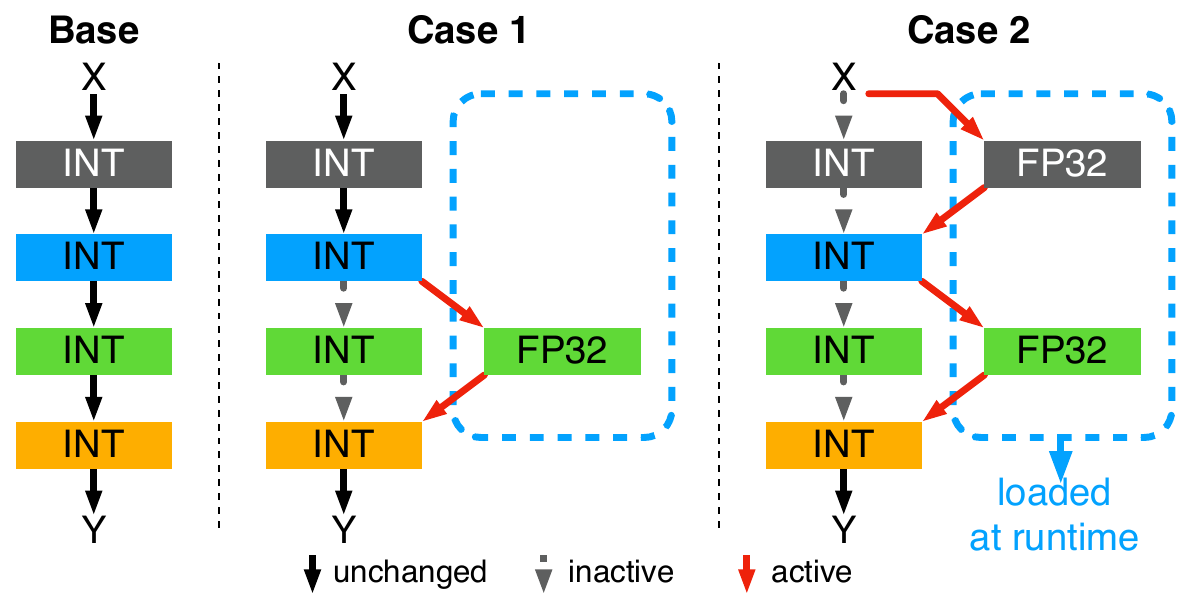} 
    \end{center}
    \vspace*{-3mm} 
    \caption{Runtime model reconfiguration and execution mechanism.}
    \label{fig:loading}
\end{figure}

\section{Dynamic Model Reconfiguration \& Execution}
\label{sec:runtime-change}
\looseness=-1
At the core of \name\ is finding the most suitable mixed-precision 
model for each app by reflecting the current system condition; 
app processes may otherwise be unstable for DNN models due
to their low accuracies or 
severely underutilize the platform resources. 
Thus, we need to consider various compressed models with different 
bit configurations, their saving and loading of many models. 
However, reloading the resulting entire model each time may 
be neither suitable for real-time apps nor efficient 
when only a few layers need to be loaded.
Thus, we propose to reconfigure the model bit-width 
configuration at runtime by lazy loading FP layers 
in order to reduce the model loading overhead 
and ensure deadline satisfaction.
This dynamic model reconfiguration approach 
is inspired by \cite{tang2022arbitrary, bulat2021bit}
where bit-width decisions are made 
only for a {\em single} DNN model, and
training a special DNN from scratch is necessary 
for runtime adaptive quantization.
In contrast, we develop a solution that determines 
bit configurations for multiple DNNs 
without requiring any additional (re)training.

\looseness=-1
In \name, when the optimal bit setting for each app is 
determined, only the additional layers for the target model
are loaded to compose the path ({\small\circled{7}} in Table~\ref{tbl:navigation}).
Before operating this at runtime, we separate a model into layers 
and store the full-precision (FP) and quantized (INT) versions of 
all layers ({\small\circled{1}}). That is, 2$L$ versions of layers 
are saved for a model consisting of $L$ layers.
Specifically, operation type (e.g., convolution) 
and FP weights are saved for an FP layer; for an INT layer, 
additional parameters required for quantization of FP activations 
are saved with operation type and INT weights.

\noindent\textbf{\textit{Mechanism.}} Model reconfiguration 
consists of 1) loading necessary layers and 2) changing the 
model execution path. Here we deal with a multi-tasking system 
in which the main process is in charge of process monitoring 
and memory allocation, and the subprocess is in charge of 
model execution.
To load only the layers needed at runtime, \name~should first 
hold the fully-quantized models of all possible apps as their 
base models by loading their INT layers and building them 
without changing the original architecture. 
When a new camera image enters to the system and any change in the status of concurrent apps is detected, the main process adjusts the amount 
of memory resource allocated to the concurrent apps 
({\small\circled{5}}).
After determining the new bit setting from the adjusted memory 
space for each app, the necessary FP 
layers required for the new model are loaded ({\small\circled{6}}), and then they 
are executed instead of the existing INT layers. 
After execution, the FP layers are deleted, and each model 
returns to its original base model state.
Fig.~\ref{fig:loading} shows the above mechanism with two 
different cases.
How to build a mixed-precision model while considering the given 
memory space will be elaborated next.

\section{System Architecture}
\label{sec:architecture}

Our main objective is to optimize the allocation of system 
resources, such as CPU time and memory space, for multiple simultaneous apps while meeting the 
deadline, which has not yet been 
addressed adequately because of the high computational 
complexity of real-time CNN models.

\looseness=-1
Representing a model using both high and low precisions 
({\small\circled{2}}) is key to reduction of the computational 
cost of DNNs, thereby enabling multiple DNN apps to share 
limited platform resources.
Central to \name\ is a memory resource allocation algorithm 
that determines the best bit setting for each app to use 
so as to maximize total accuracy ({\small\circled{5})}.
This algorithm uses app profiles that are created offline to 
save the information of a set of mixed-precision models 
({\small\circled{3}}), as a constraint term. 
Thus, optimization of computation resource is triggered
by the apps whose memory requirements change dyamically.
After completing a series of processes for resource optimization, 
the additional layers to load/execute are selected by 
looking up tables ({\small\circled{6}}), 
which were generated offline ({\small\circled{4}}).

To reduce the online burden of computation-heavy processes, 
\name\ is divided into offline and online stages (see 
Fig.~\ref{fig:overview}).
{\small\circled{1}, \small\circled{7}} are discussed in 
Sec.~\ref{sec:runtime-change}, and the remaining components 
are elaborated in the subsequent sections.

\begin{figure}[tp]
    \begin{center}
    \includegraphics[width=0.9\linewidth]{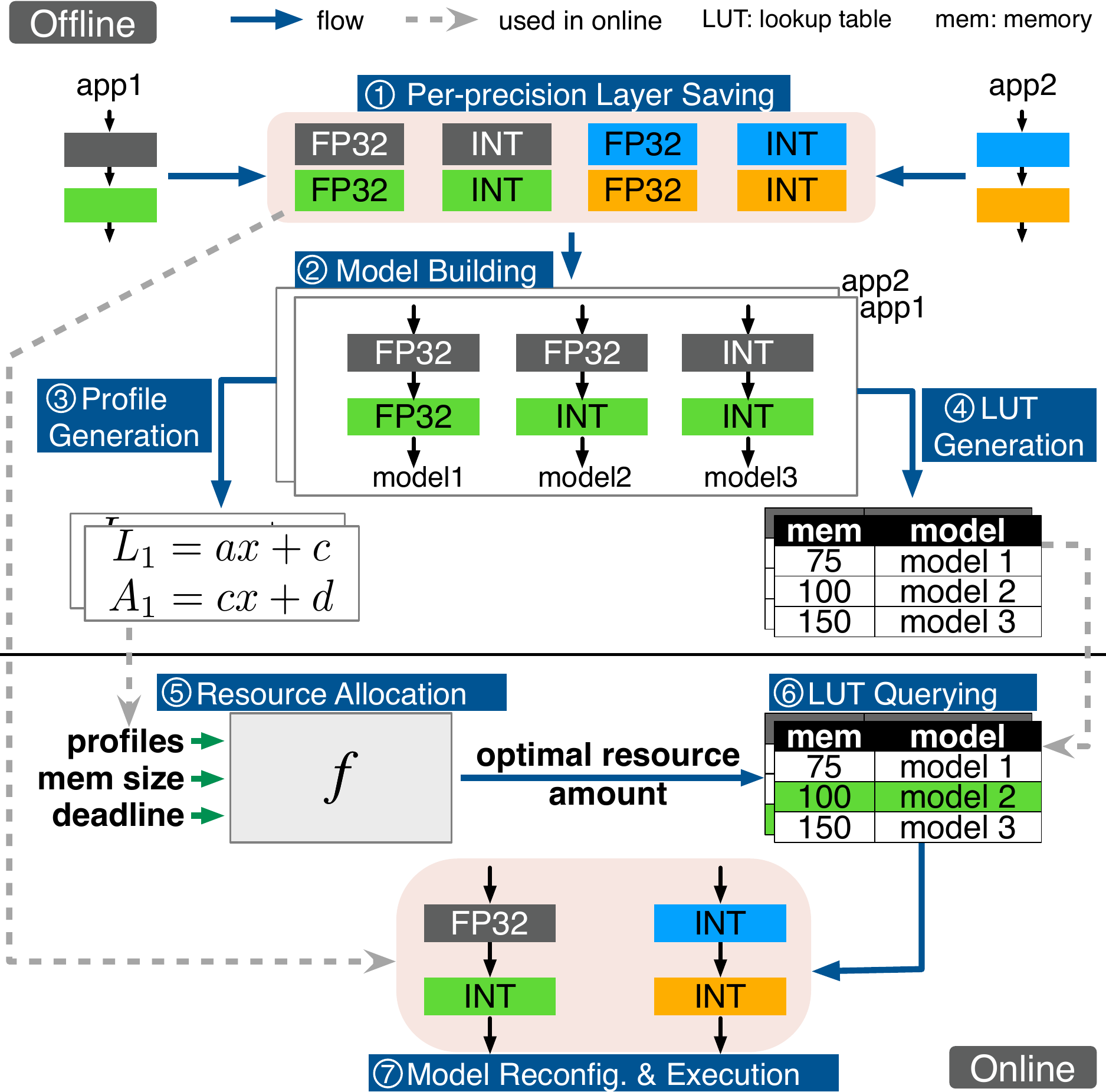} 
    \end{center}
    \vspace*{-3mm} 
    \caption{System architecture.}
    \label{fig:overview}
    \vspace*{-2mm} 
\end{figure}

\begingroup
\setlength{\tabcolsep}{6pt}
\renewcommand{\arraystretch}{1.0}
\begin{table}[t]
    \caption{Navigation for system components. PIC means the process in charge of the corresponding component.}
    \begin{center}
    \begin{tabular}{clcc}
        \hlineB{2}
        \textbf{Stage} & \multicolumn{1}{c}{\textbf{Component}} & \textbf{Section} &\textbf{PIC}\\
        \hlineB{2}
        \multirow{4}{*}{Offline}
        &{\small\circled{1}} Per-precision Layer Saving & \ref{sec:runtime-change} & Sub\\ \cline{2-4}
        &{\small\circled{2}} Mixed-precision Model Building & \ref{subsec:compression} & Sub\\ \cline{2-4}
        &{\small\circled{3}} Profile Generation & \ref{subsec:profiling} & Main\\ \cline{2-4}
        &{\small\circled{4}} Lookup Table Generation & \ref{subsec:lookup} & Main\\
        \hlineB{1.5}
        \multirow{3}{*}{Online}
        &{\small\circled{5}} Memory Resource Allocation & \ref{subsec:allocation} & Main\\ \cline{2-4}
        &{\small\circled{6}} Lookup Table Querying& \ref{subsec:lookup} & Main\\ \cline{2-4}
        &{\small\circled{7}}  Model Reconfig. and Execution & \ref{sec:runtime-change} & Sub \\
        \hlineB{2}
    \end{tabular}
    \end{center}
    \label{tbl:navigation}
    \vspace*{-3mm} 
\end{table}
\endgroup

\subsection{Mixed-Precision Model Building} 
\label{subsec:compression}
\looseness=-1
Using the FP and INT versions of all layers, we can build 
different mixed-precision models (based on the identical 
architecture) which show the effects of mixed-precision 
quantization in terms of accuracy, speed, model size, etc.
The main objective of \name~is to find a 
model that maximizes the accuracy with the given resources 
and deadline.
We generate different compressed models of 
various bit configurations (showing different performances), 
and use them for the prediction of performance in accordance 
with the amount of available memory resource.

We develop a sorting-based approach to build models while yielding 
the highest accuracy compared to the degree of compression
(defined as the number of the quantized layers in the model). 
In this approach, the FP layers of the model are 
sorted in descending order of layer sensitivity discussed in 
Sec.~\ref{sec:sensitivity}.
ZeroQ \cite{cai2020zeroq} shows the overall sensitivity of a quantized 
model (where multiple layers are quantized with low bit-widths) 
can be calculated by summing up the sensitivities of all 
quantized layers in that model.
Based on this calculation, starting with the fully-quantized 
model, the INT layers are replaced with their corresponding 
FP layers one-by-one (in descending order of layer sensitivity). 
This means that at all degrees of quantization, the generated 
model would yield the smallest output perturbation from the 
full-precision model, thus leading to the smallest accuracy drop.
This model building process is described in 
Algorithm~\ref{alg:quantization}.
Although this can be presented recursively or iteratively like 
depth-first search, we present it recursively for readability,
and the initial invocation for this procedure is 
\textsc{build}~(Fully-quantized model, $\emptyset$).

Our sorting-based approach yields $L$ compressed models for a 
model consisting of $L$ layers, which effectively removes 
unnecessary and repetitive models among $2^L$ models produced by 
an exhaustive model generation approach.

\looseness=-1
Fig.~\ref{fig:profiles} shows the inference results of the 
compressed models of VGG16 \cite{simonyan2014very}, 
ResNet50 \cite{he2016deep}, and YOLO-v3 \cite{redmon2016you}.
Note latency includes the time for model reconfiguration, 
execution, and restoration.
We used 2 bit-widths, FP32 and INT8, throughout this paper to 
demonstrate the effectiveness of using mixed-precision models 
on app performance, i.e., all layers are
expressed with either precision.
The three graphs in the upper row show latency changes and 
the three graphs in the lower row display accuracy changes as 
a function of the degree of compression.
Considering memory usage and latency of a model may change each time 
we run the model, we measured 100 times for each model and used the 
highest peak memory usage and latency as the worst case.
The results show a latency reduction without sacrificing accuracy 
much when compression degree is increased.
To utilize the performance results of the compressed models in the 
online stage, two profiles and a lookup table are created for 
each model.

\begin{algorithm} [t]
    \caption{Mixed-Precision Model Building} 
    \begin{algorithmic}
    \State $L$  \textbf{:} a set of layers sorted in descending order by 
        layer sensitivity
    \State $\mathcal{S}$  \textbf{:} a set of compressed models ($\mathcal{S} 
       \gets \emptyset$)
      \\
    \Procedure{Build}{model $m$, layer $l$}
        \State $n$ = Replace $l$ of $m$ with FP layer\Comment{\small If $l$ is $\emptyset$, $n \gets m$}
        \State $\mathcal{S} \gets \mathcal{S}\cup n$
        \State $v$: the most sensitive layer in $L$\Comment{\small If $L$ is $\emptyset$, \Return {$\mathcal{S}$}}
        \State remove $v$ in $L$
        \State \textsc{BUILD}($n$, $v$)
        \State \Return {$\mathcal{S}$}
    \EndProcedure
    \end{algorithmic}
    \label{alg:quantization}
\end{algorithm}

\subsection{Profile Generation}
\label{subsec:profiling}
The above process yields a number of mixed-precision models for 
each app, and hence the goal of our memory resource allocation is 
to find 
an optimal set of compressed models for all concurrent apps.
Although such a large number of combinations offers
flexibility to accommodate various real-world cases, the high 
computational cost of finding an optimal set makes it 
unsuitable for real-time resource allocation. 
Our model compression yields $O(L)$ models, and in the case of 
a 10 FPS camera, the end-to-end latency per frame should be 
less than 100 ms \cite{yang2019re}.
Given that $I$ apps are simultaneously running on the platform, 
an excessive amount of time (up to $O(100 \cdot L^{I})~ms$) is 
required to find an optimal combination.

\begin{figure}[tp]
    \begin{center}
    \includegraphics[width=1.0\linewidth]{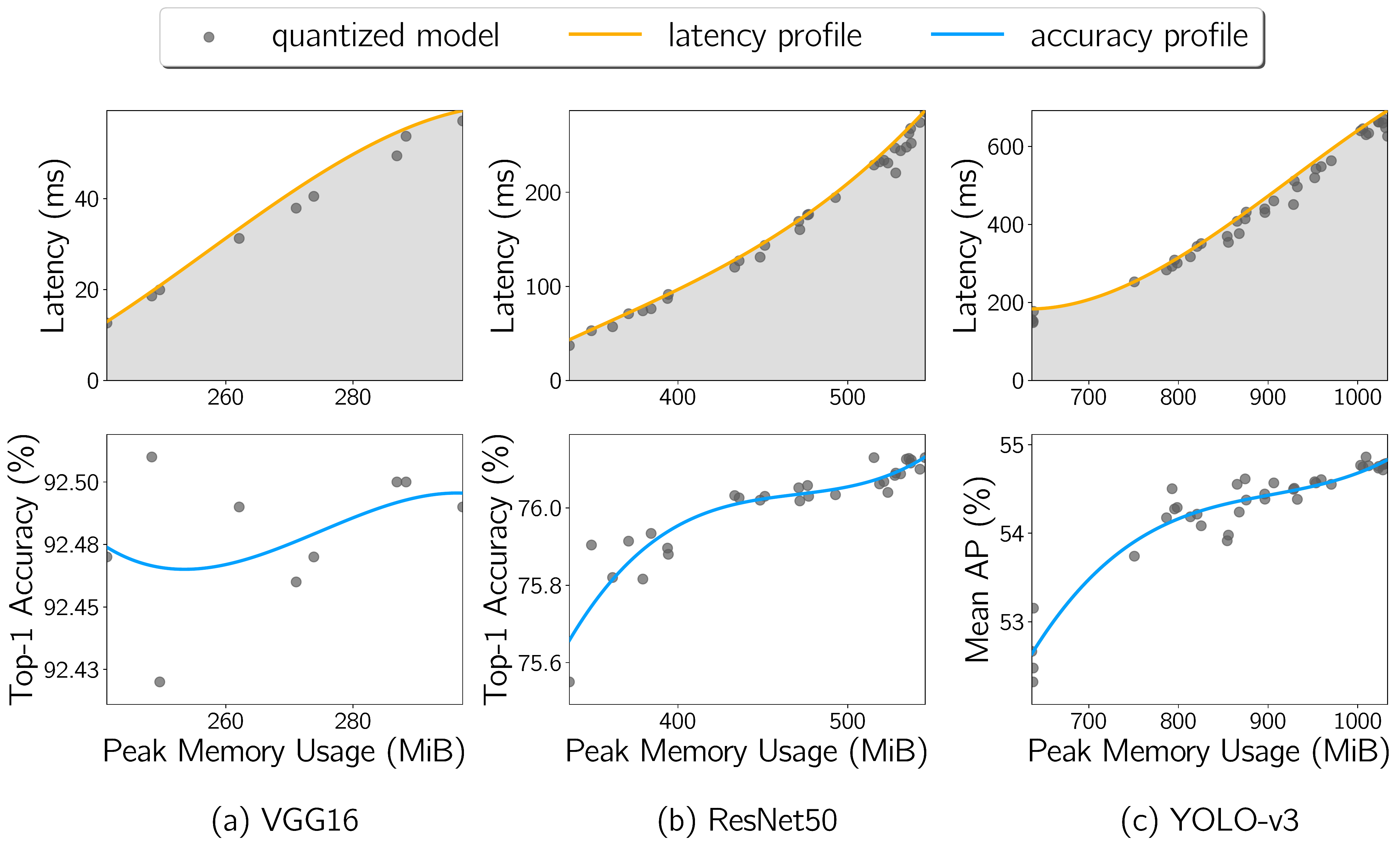} 
    \end{center}
    \vspace*{-3mm} 

    \caption{The results of model compression and profiling on 
    VGG16, ResNet50, and YOLO-v3. Each gray dot indicates a 
    certain compressed model that has a specific bit setting. 
    A yellow line denotes the profile of worst-case latency, 
    thus covering all the compressed models as shown in the gray 
    areas. A blue line means the profile of 
    most-likely accuracy, showing the trend of accuracy.}
    \label{fig:profiles}
    \vspace*{-2mm} 
\end{figure}

\begingroup
\setlength{\tabcolsep}{10pt}
\renewcommand{\arraystretch}{1.3}
\begin{table}[tp]
\begin{centering}
\caption{Latency and accuracy profiles. X means peak memory usage.}

\vspace*{-2mm} 
\begin{tabular}{>{\centering\arraybackslash}m{1.4cm} >{\centering\arraybackslash}m{1cm} >{\centering\arraybackslash}m{4.3cm}}
\hlineB{2}
\textbf{NN Model}         & \textbf{Type} & \multicolumn{1}{c}{\textbf{Profile}}                                               \\ \hlineB{2}
\multirow{2}{*}{VGG16}    & Latency       & $-0.0001472~x^3 + 0.1135 x~x^2 - 28.16 ~x + 2267$   \\[0ex] \cline{2-3} 
                          & Accuracy      & $-7.781e\mhyphen07~x^3 + 0.0006417~x^2 - 0.1753~x + 108.4$ \\[0ex] \hline
\multirow{2}{*}{ResNet50} & Latency       & $1.021e\mhyphen05~x^3 - 0.01083 ~x^2 + 4.648~x - 683$     \\ \cline{2-3} 
                          & Accuracy      & $1.285e\mhyphen07~x^3 - 0.000181~x^2 + 0.08553~x + 62.48$ \\ \hline
\multirow{2}{*}{YOLO-v3}  & Latency       & $-6.706e\mhyphen06~x^3 + 0.01861~x^2 - 15.5~x + 4241$      \\ \cline{2-3} 
                          & Accuracy      & $6.764e\mhyphen08~x^3 - 0.0001834~x^2 + 0.1676~x + 2.821$ \\ \hlineB{2}
\end{tabular}
\vspace*{-3mm} 
\label{tbl:profiles}
\end{centering}
\end{table}
\endgroup

For real-time optimization, the authors of NestDNN \cite{fang2018nestdnn} 
considered only five candidate models, which come at the cost of 
both resource utilization and accuracy. 
In \name, however, profiles are generated offline for each app 
in the form of polynomials for latency and accuracy by 
applying polynomial regression to the compressed models.
This app performance profiling eliminates the need 
of additional inference processing in the online stage. 
We introduce latency and accuracy profiles, which are used 
in our online memory resource allocation algorithm.

\subsubsection{Latency Profile}
By employing the latency profile of a target DNN, we can estimate the
worst-case latency when the model is compressed to a certain size. 
If some models cannot be covered by the latency profile, the newly 
selected model at runtime can exceed the expected number obtained 
from the latency profile, thus failing to guarantee the deadline. 
Thus, the latency profile should cover all the resulting 
compressed models. The impact of inter-task interference in a 
multi-tasking platform on execution latency is discussed in 
Sec.~\ref{sec:eval_feasibility}.

To create the latency profile that covers all compressed models, 
the compressed models are stored when its latency is 
greater than the latencies of the smaller models, and then 
polynomial regression is applied to the stored models.
We exploit cubic polynomials because latency is not completely 
but fairly proportional to peak memory usage, and high-order 
regression burdens the optimization algorithm (see 
Fig.~\ref{fig:profiles}).
After obtaining a polynomial profile for each model, 
to guarantee the worst-case latency for a given memory space,
we delete those models whose latencies are greater 
than the values estimated 
with the latency profile from a set of all compressed models.
The resulting latency profiles of VGG16, ResNet50, and YOLO-v3 
are provided in Table~\ref{tbl:profiles}.

\subsubsection{Accuracy Profile}
The role of the accuracy profile is different from that of the 
latency profile where we observe the change in accuracy according 
to the degree of compression. For this, we apply regression 
to all of the remaining models after creating the latency profile, 
without deleting models any further.
Accuracy also has a weak positive linear correlation with peak 
memory usage, and thus regression is performed using a cubic 
polynomial (see Fig.~\ref{fig:profiles}). The accuracy profiles 
of VGG16, ResNet50, and YOLO-v3 are provided in 
Table~\ref{tbl:profiles}. 

\subsection{Optimal Resource Allocation}
\label{subsec:allocation}
Once a new camera image enters to the system and any change 
in the status of app processes is detected, 
\name\ allocates the memory resource for concurrent apps 
while maximizing the total accuracy under the given deadline.
The optimization problem in this process is solved by reflecting 
the latency and accuracy profiles generated in the offline 
stage to determine the optimal allocation of memory resource 
for each app. In what follows, we present the development of 
a resource allocation algorithm, which is formulated as a 
constrained nonlinear optimization problem.

\name\ assumes a limited memory capacity, and the timing 
requirement of a real-time app is determined
by camera's frame rate.
Therefore, the problem of resource 
allocation for concurrent apps is formulated as: 

\begin{equation}
\begin{aligned}
\label{eqn:rsc_optimization}
& \arg \max_{m_{i}} && \sum_{i \in I} \lambda_{i}A_{i}\left(m_{i}\right) \\ 
& \textrm{s.t.} && \sum_{i \in I} L_{i}\left(m_{i}\right) \leq D - \varepsilon \\
&&& {\forall{i} \in I}:m_{i} \leq M_{\max } - \mu_{i} \\
&&& {\forall{i} \in I}:m_{i}^{L} \leq m_{i} \leq m_{i}^{U}
\end{aligned}
\end{equation}

\looseness=-1
\noindent
where $I$ is the set of concurrent apps. 
We want to find the optimal memory size
$m_{i}$ for each app $i \in I$ that maximizes total accuracy.
{\em Total accuracy} is calculated as the weighted sum of 
accuracies of all apps, and $\lambda_{i}$ is used to give different 
weights when each app has a different level of importance or the 
degree of change in accuracy during quantization varies with app.
$\lambda_{i}$ is set to 1 in this paper.
For the constraint terms, we consider memory capacity apps can 
use $M_{\max}$, and the given deadline $D$. 
Here, $\varepsilon$ is the average processing time in the main 
process, and $\mu_{i}$ is the memory usage of the base models 
of concurrent apps, which is calculated as 
$\mu_{total}-\mu_{app_{i}}$, where  $\mu_{total}$ is $\mu_{vgg}+\mu_{resnet}+\mu_{yolo}$, and
$\mu_{app_{i}}$ is one of 
$\mu_{vgg}$, $\mu_{resnet}$, and $\mu_{yolo}$.
$\varepsilon$ and $\mu_{i}$ will be detailed in Sec.~\ref{sec:eval_feasibility}
in which we assess the feasibility of solving this problem.
We also use the latency profile $L_{i}$ and accuracy profile 
$A_{i}$ obtained from the profile-generation phase. 
The latency-related constraint term ensures that estimated latency 
$L_{i}\left(m_{i}\right)$ does not exceed the deadline. 
Lastly, we set lower and upper bounds ($m_{i}^{L}$, $m_{i}^{U}$) for 
$m_{i}$ to reduce the search space in solving this optimization 
problem. $m_{i}^{L}$ is set to the smallest peak memory usage of 
app $i$, and $m_{i}^{U}$ is set to the largest peak memory usage 
of app $i$. The optimization result determines the appropriate 
memory space size to be allotted to each app. 

\begin{algorithm} [t]
    \caption{Lookup Table Generation for App $i$} 
    \begin{algorithmic}
    \State $\mathcal{T}$  \textbf{:} lookup table ($\mathcal{T} \gets \emptyset$)
    \State $\mathcal{M}$ \textbf{:} a set of mixed-precision models
    \State \textbf{$z$}  \textbf{:} The number of lookup table entries
    \State \textbf{$max(i)$}  \textbf{:} the largest peak memory usage
    \State \textbf{$min(i)$}  \textbf{:} the smallest peak memory usage
    \State \textbf{$m_{pm}$}  \textbf{:} peak memory usage of model $m$
    \State \textbf{$m_{acc}$}  \textbf{:} accuracy of model $m$
    \\ 
    \State  Sort $\mathcal{M}$ in ascending order by peak memory usage
    \State $val \gets 0, oldkey \gets 0$
    \While {$m \in \mathcal{M}$}
        \State $ key = min(i)
                        + \Bigl\lceil{ \frac{(z-1)(m_{pm} - min(i))}{max(i)-min(i)} \Bigr\rceil}
                        \frac{max(i) - min(i)}{(z-1)} $
                        
        \If{$key == oldkey$}
            \If{$m_{acc}$ is greater than $val_{acc}$}
                \State $val \gets m$
            \EndIf
        \Else
            \State $\mathcal{T} \gets \mathcal{T} \cup (oldkey,  val)$
            \State $val \gets m$, $oldkey \gets key$
        \EndIf
        \State remove $m$ in $\mathcal{M}$
    \EndWhile
    \end{algorithmic}
    \label{alg:lookup}
\end{algorithm}

\subsection{Lookup Table for Final Model Selection}
\label{subsec:lookup}

After resource allocation, \name~should determine which 
compressed models to load into the system.
Given a large number of compressed models for each DNN model, 
searching for the best fitting model sequentially will likely 
take too long for real-time apps.
We thus reduce this search time via lookup tables, which are 
generated for each model in the 
offline stage and used in the online stage.

\subsubsection{Generation of Lookup Table}
\looseness=-1
We present a simple yet efficient method for generating the lookup 
table, which is based on the principle of arithmetic sequences.
We first divide equally the range of peak memory usage of each app.
Second, a new entry is added to the table in the form of a key--value 
pair, where the key is the endpoint of a certain divided section, 
and the value is the bit setting of the model with the highest 
accuracy when the memory size corresponding to that end-point 
is allocated.
Starting from the smallest section, the entry for each section 
is iteratively inserted into the table.
As a result, the resulting lookup tables remain to be of constant 
size (i.e., the preset number of table entries) regardless of the 
complexity of the DNN model.
Fig.~\ref{fig:lookup_table} shows the accuracy of the models of 
which bit settings are stored in the lookup table.
Each graph shows the number of different settings \name\ needs 
to store for each app (e.g., 2 for VGG16). The detailed procedure 
for lookup table generation is described in 
Algorithm~\ref{alg:lookup}.

\begin{figure}[pt]
    \begin{center}
     \includegraphics[width=1\linewidth]{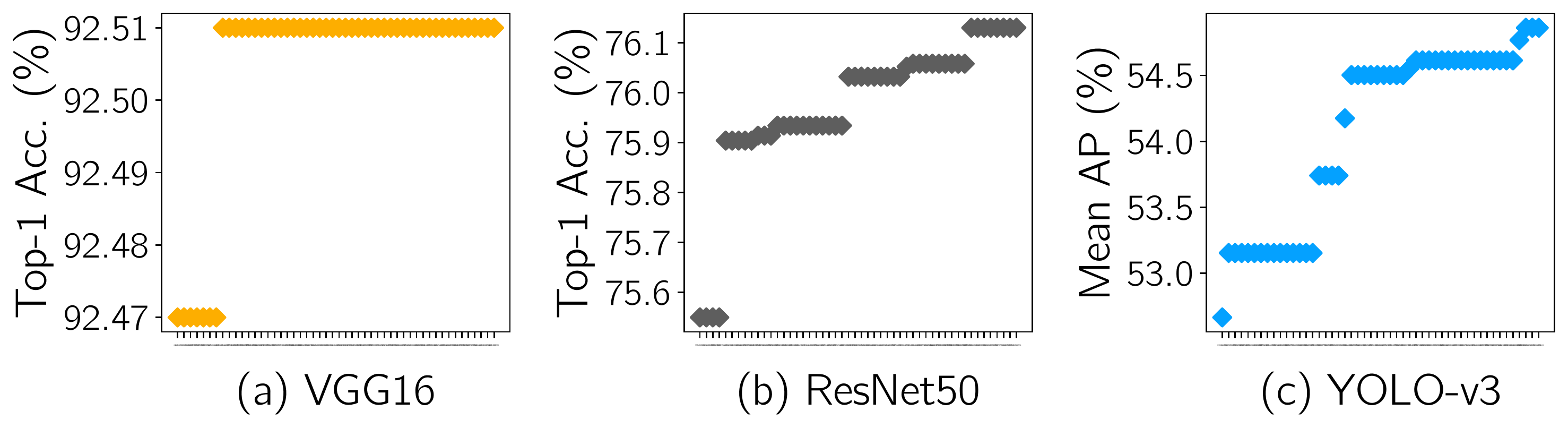} 
    \end{center}
    \vspace*{-5mm} 
    \caption{The accuracies of the models stored in the lookup table. 
    The ticks marked on each x-axis denote the sections created by dividing 
    the range of peak memory usage by a fixed number (i.e., the number of 
    lookup table entries). Each of the resulting lookup tables contains 
    50 entries.}
    \label{fig:lookup_table}
    \vspace*{-5mm} 
\end{figure}

\subsubsection{Querying Lookup Tables}
The generated lookup tables allow \name~to select the final model 
quickly after the online resource allocation.

\begin{align*}
{
    key = min(i) + \Bigl\lfloor{ \frac{(z-1)(m_{i}^{\star} - min(i))}{max(i)-min(i)} \Bigr\rfloor}\frac{max(i) - min(i)}{(z-1)}
}
\end{align*}
where $m_{i}^{\star}$ denotes the memory size allocated for 
app $i$, and $z$ is the number of table entries. 
$max(i)$ and $min(i)$ represent the maximum and minimum peak 
memory usages of app $i$. 

After entering the resulting keys into the tables, the final models 
to execute are determined. 
Each subprocess of an app then loads the additional FP 
layers required for the newly selected model and runs 
according to the reconfigured execution path as discussed in 
Sec.~\ref{sec:runtime-change}.

\section{Evaluation}
\label{sec:evaluation}

We now evaluate \name\ by implementing all the components 
in Table~\ref{tbl:navigation} in a simulated environment.
We implemented layer quantization and model execution 
using PyTorch 1.10.2 \cite{pytorch}, which is one of the most 
representative deep learning frameworks.
In compliance with the quantization support by 
PyTorch, INT8 is used as low precision on 
x86-64 CPU for the acceleration of DNN execution.
Instead of using indirect metrics for app performance, 
we measured actual latency by running all the compressed models 
and peak memory usage by using memory profiler \cite{mem_profiler}.
For profile generation, we used the polynomial regression 
utility, polyfit, in the NumPy API.
The resource allocation algorithm was programmed using the 
sequential least square programming utility in SciPy API.
All our experiments were conducted on an AMD Ryzen 7 
5700G processor and 16GB RAM with Ubuntu Linux 20.04 
LTS operating system.
Given the goal of \name, we want to meet the same
deadline of concurrent DNN apps.

\looseness=-1
Following \cite{heo2020real, kang2022dnn}, the target deadline 
is relatively determined by reflecting the performance of 
the original (i.e., full-precision) models measured in our environment.
Even though we experimented in a simulated environment, 
we aim to show \name’s ability of resource optimization
when the state of the running apps changes.


\noindent\textbf{\textit{Multi-tasking Implementation.}}
Our multi-tasking environment consists of the main 
and sub-processes. The main process detects changes in the system,
and allocates platform resources to concurrent apps.
Each subprocess is responsible for the execution of its DNN model. 
Note that the number of subprocesses 
is equal to that of types of possible apps. 
For multi-tasking, we use the Non-Preemptive Shortest Job 
First (SJF) scheduling to avoid the high overhead of context 
switching between DNN-based apps, consider the equal 
priority between apps (attributed to the common deadline), 
and minimize the average waiting time of concurrent apps. 
For example, the three apps introduced throughout 
this paper are scheduled in the order of VGG16, ResNet50, 
and YOLO-v3.

As discussed in Sec.~\ref{sec:runtime-change}, each subprocess 
first builds the fully-quantized model of its app as a base model.
When processing images coming from the camera, 
\name\ behaves differently depending on whether any change 
in the status of concurrent apps is detected or not.
In case a change occurs, the main process determines the optimal 
memory space for each app, and then notifies the corresponding bit 
configuration to one of the subprocesses by sending a message 
(\textbf{\em case~1}). 
Otherwise, the main process does not make resource allocation 
and notifies the last bit configuration to the subprocess 
(\textbf{\em case~2}).
When the subprocess receives the information, it loads the 
necessary layers, runs the model, and then reports to the main 
process that the job is over through shared memory.
Then, the main process 
repeats this procedure until all apps are executed.
In the experiments in Sec.~\ref{sec:scenario_test}, we adjusted 
the status of apps at 1s intervals.
If \textbf{\em case~1} meets the deadline, so does \textbf{\em case~2}.
We therefore show the results of deadline satisfaction 
for \textbf{\em case~1}.

In what follows, we first present the DNN models and datasets 
used in our evaluation, and discuss the feasibility, effectiveness,  
and robustness of \name.

\begin{figure*}[pt]
    \begin{center}
    \includegraphics[width=0.9\linewidth]{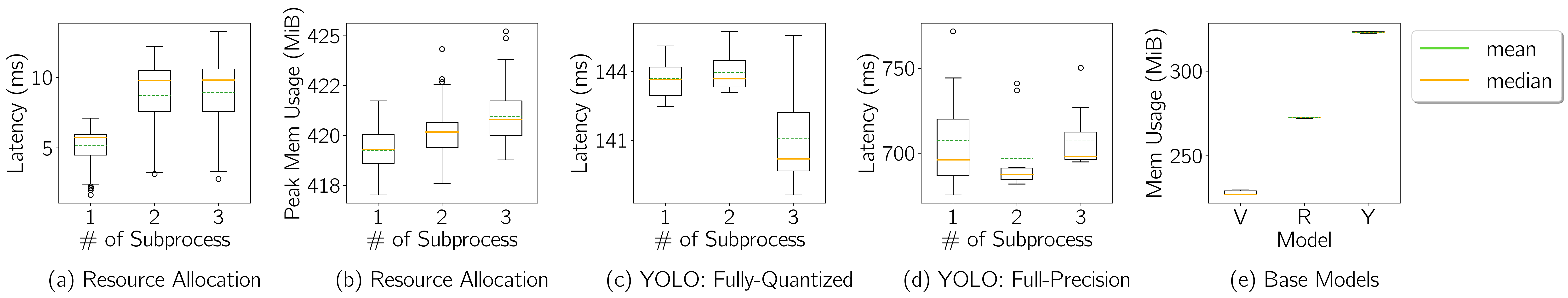} 
    \end{center}
    \vspace*{-4mm} 
    \caption{Distributions of the latency and peak memory usage of 
    main and sub-process. 
    If the number of subprocesses is 1, YOLO-v3 is loaded and 
    executed. In case of 2, ResNet50 and YOLO-v3 are loaded and 
    executed, and in case of 3, all three apps are loaded and 
    executed. In the last graph, `V', `R', and `Y' stand for 
    `VGG16', `ResNet50', and `YOLO-v3', respectively.
    Each box contains the results of 100 measurements.}
    \label{fig:mainproc}
    \vspace*{-4mm} 
\end{figure*}

\subsection{Neural Networks and Datasets}
\label{sec:eval_networks}

Image classification and object detection apps often 
run simultaneously on AVs.
Furthermore, even for identical tasks, the complexity of model 
varies greatly with the function.
For example, classifying objects in front of a car is more 
difficult than classifying the color of a traffic light.
Considering the diverse real-world cases, the following three 
DNN models were selected for two types of apps.
Although the state-of-the-art models change rapidly, their 
underlying architectures are predicated on the models of 
our choice \cite{yun2019cutmix, ding2021repvgg}.

\subsubsection{VGG16 on CIFAR-10}
CIFAR-10 \cite{cifar10} is made up of 60K images in 10 classes, 
consisting of a low-resolution color image of animals or vehicles.
Classification of CIFAR-10 is considered to be a simple and easy 
problem owing to the small number of classes and the clear visual 
distinction between class objects.
Based on CIFAR-10, we used VGG16,\footnote{Obtained from 
https://github.com/chengyangfu/pytorch-vgg-cifar10} which is 
fully trained on the dataset.
VGG is one of the most popular and representative architectures \cite{lee2021novel, zhe2019optimizing, fang2018nestdnn}, which 
uses small (e.g., 3$\times$3) kernels in all convolution layers.
In particular, VGG16 (consisting of 16 layers) shows nearly 
state-of-the-art inference accuracy on the CIFAR-10 dataset.
As a result of the simplicity of the task, VGG16 shows higher 
accuracy than the other models as shown in 
Figs.~\ref{fig:profiles} and~\ref{fig:lookup_table}.

\subsubsection{ResNet50 on ImageNet}
ImageNet \cite{ILSVRC15} is composed of 1.4M images in 1K classes, 
each of which is a high-resolution color image of animals, plants, 
vehicles, or electronic devices.
Classification on ImageNet is more difficult than on CIFAR-10 
because of the high similarity of features between classes and 
a large number of classes.
The ResNet50 model,\footnote{Obtained from 
https://github.com/pytorch/vision/blob/main/torchvision/} 
fully trained on the ImageNet dataset, was used in this study.
ResNet, released after VGG, has a much deeper neural network 
architecture (a depth of up to 152 layers) than VGG (a depth of 
up to 19 layers), which solves the vanishing gradient problem 
caused when designing deep networks by adding skip connections 
to the network. 
Due to the complexity of ImageNet, the accuracy of ResNet-50 is lower 
than that of VGG16 on CIFAR-10 (see Figs.~\ref{fig:profiles} 
and~\ref{fig:lookup_table}), but currently widely-used ImageNet-based 
models show accuracies of 75 $\sim$ 80\%, and ResNet-50 shows 
nearly state-of-the-art inference accuracy on ImageNet.
More importantly, the ResNet architecture has great potential for 
building the current high-performance deep model \cite{touvron2019fixing}.

\subsubsection{YOLO-v3 on MS COCO}

For the object detection task, we used the YOLO-v3 
model\footnote{Obtained from 
https://github.com/eriklindernoren/PyTorch-YOLOv3} trained on the MS 
COCO dataset \cite{lin2014microsoft}, which is composed of 5K images 
in 91 classes, consisting of images pertaining to our daily life 
such as one or more objects of vehicles and daily necessities.
YOLO is the first object detection architecture that localizes and 
classifies multiple objects in an input image or video in one single 
forward propagation. Thanks to its speed, RTOD has been realized 
in autonomous driving \cite{yang2019re}, and its role in promoting 
safety and convenience continues to grow.
In Figs.~\ref{fig:profiles} and~\ref{fig:lookup_table}, the accuracy 
of YOLO-v3 (measured in ``average precision (AP)'') is lower than 
that of the classification models (measured in ``Top-1'').
This is because AP means not only the accuracy of classification
but also the prediction accuracy of each object position, 
whereas Top-1 only indicates classification accuracy. 
However, YOLO-v3 performs well in balancing accuracy and latency and 
is one of the most commonly used RTOD models in both academia 
and industry \cite{yang2019re, lin2018architectural, kangdnn}.

\subsection{Feasibility of Resource Allocation Algorithm in \name}
\label{sec:eval_feasibility}
Ensuring the feasibility of resource allocation depends
on the following factors: 
1) the overhead and additional memory usage in solving 
Eq.~(\ref{eqn:rsc_optimization}), 2) the reasonableness 
of using latency profiles for the first constraint term  
of Eq.~(\ref{eqn:rsc_optimization}), and 3) the memory 
usage of each app’s base model for the second constraint 
term of Eq.~(\ref{eqn:rsc_optimization}). 
We show the feasibility of the 
proposed algorithm by identifying each of these factors.

\subsubsection{Runtime overhead and memory usage in 
solving Eq.~(\ref{eqn:rsc_optimization})} 
Eq.~(\ref{eqn:rsc_optimization}) entails a cubic 
objective function with cubic and linear constraints. 
Although the size of the search space is reduced by using 
lower/upper bounds, we need to assess its runtime overhead 
to ensure the deadline satisfaction.
Fig.~\ref{fig:mainproc}~(a) shows the number of concurrent 
apps vs.~the latency of solving 
Eq.~(\ref{eqn:rsc_optimization}).
Latency in the single-tasking case was about 
1.7$\times$ smaller than the multi-tasking case.
Based on this observation, we set $\varepsilon$
in Eq.~(\ref{eqn:rsc_optimization}) to 15 ms by considering
the maximum latency of resource allocation.
Meanwhile, Fig.~\ref{fig:mainproc}~(b) shows the peak 
memory usage is found almost constant regardless of 
the number of running apps (the ratio of the smallest
to the largest mean is approximately 1).
So, given that $M_{max}$ is the memory capacity 
for the subprocesses of the apps, not for the main process,
the main process does not require more space and hence
has little effect on $M_{max}$.

For the various cases in real-world,
the latency and peak memory usage values (for each box) in 
Figs.~\ref{fig:mainproc}~(a) and (b) are measured while 
changing deadline and memory capacity.

\subsubsection{Reasonableness of using latency profiles}
Given the latency profile is based on the measurements in isolation, 
the latencies should hold even when multiplexed with other apps.
From Figs.~\ref{fig:mainproc}~(c) and (d), one can observe 
that the latency of each compressed model holds in the non-preemptive 
SJF scheduling-based multi-tasking environment.
Due to space limitation, we only showed two models: 
(c) for the fully-quantized model of YOLO-v3, 
and (d) for its full-precision model.
From these, we conclude that the latency profile can be used 
for the first constraint term.

\subsubsection{Memory usage of each app’s base model}
\looseness=-1
More concurrent apps occupy more memory space by loading 
their base models.
Fig.~\ref{fig:mainproc}~(e) shows the memory used by 
each base model. Based on their mean values, we set 
$\mu_{vgg}$, $\mu_{resnet}$, 
$\mu_{yolo}$, $\mu_{total}$ to 228, 273, 323, 823,
respectively, in the second constraint term.

\begin{figure*}[tp]
    \begin{center}
    \includegraphics[width=0.9\linewidth]{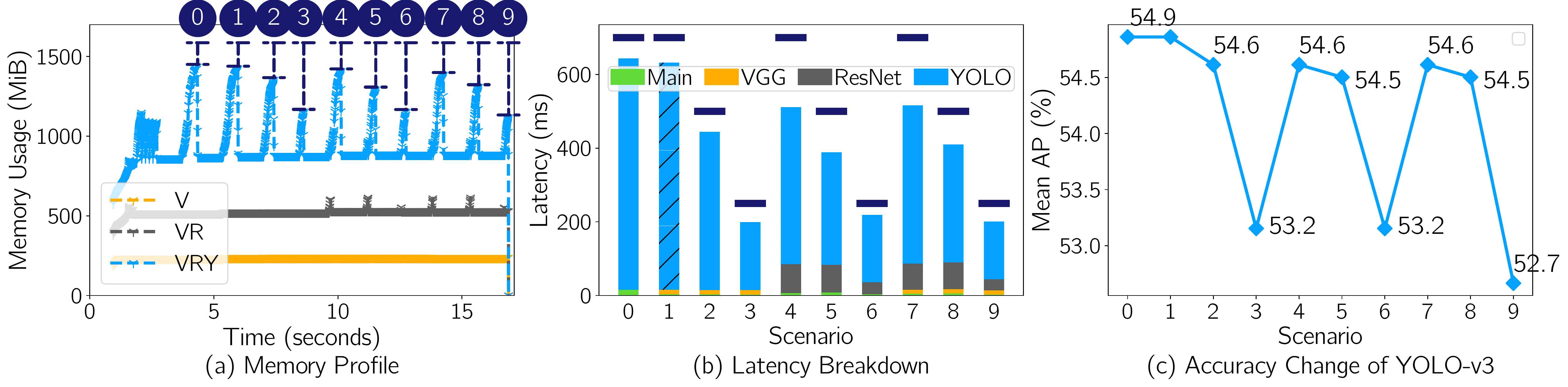} 
    \end{center}
    \vspace*{-5mm} 
    \caption{Resource optimization results. In (a), we marked the 
    margin between peak memory usage by 3 apps and the memory 
    capacity with a blue line; `V' is VGG16, `VR' is the sum of memory 
    used by VGG16 and ResNet50, and `VRY' denotes the total memory 
    space used by all 3 apps. 
    (b) shows the total latency (consisting of model reconfiguration, 
    app execution, etc.) and each deadline (marked in a blue line), 
    and (c) shows the accuracy change of YOLO-v3.}
    \label{fig:multi}
    \vspace*{-4mm} 
\end{figure*}

\subsection{Effectiveness of \name}
\label{sec:scenario_test}
\looseness=-1
To show the efficacy of \name\ in a multi-tasking 
environment, we designed scenarios where the number of concurrent apps 
changes under two types of timing constraints: sufficient and tightened 
deadlines to run multiple DNN apps (see Table~\ref{tbl:scenarios}).
Owing to the space limitation, we selected some deterministic 
yet representative (instead of random) cases where deadline 
or the number of concurrent apps is adjusted 
to show the change of each app performance.

In general, the model execution speed depends strongly 
on the processing engine (e.g., CPU, GPU).
Considering the use of much slower CPU than GPU, we set
the deadlines based on the latencies measured in our 
simulated environment. In particular, we used
minimum (148 ms), median (470 ms), and maximum (702 ms) 
worst-case latencies resulting from YOLO-v3's 
mixed-precision models.
Memory capacity was set to 1600 MiB by considering the maximum 
peak memory usage of those models.
Figs.~\ref{fig:multi}~(a) and (b) show that \name's 
resource optimization enables deadline satisfaction.

\begingroup
    \setlength{\tabcolsep}{10pt}
    \renewcommand{\arraystretch}{1.0}
    \begin{table}[tp]
        \caption{Evaluation scenarios on a multi-tasking system}
        \begin{tabular}{C{0.7cm}L{6.3cm}}
        \hlineB{2}
        \textbf{Scenario} & \multicolumn{1}{c}{\textbf{Description}}                                          \\ \hlineB{2}
        Initial          & YOLO-v3-based app is working alone ({\small{\fullsmallcircled{0}{midnightblue}{midnightblue}{white}}}).                                        \\ \hline
        1                 & VGG16-based app starts to run in {\small{\fullsmallcircled{0}{midnightblue}{midnightblue}{white}}}.                     \\ \hline
        2                 & Deadline decreases from 700 ms to 500 ms in {\small{\fullsmallcircled{1}{midnightblue}{midnightblue}{white}}}.  \\ \hline
        3                 & Deadline decreases from 500 ms to 250 ms in {\small{\fullsmallcircled{2}{midnightblue}{midnightblue}{white}}}.   \\ \hline
        4                 & ResNet50-based app starts to run in {\small{\fullsmallcircled{0}{midnightblue}{midnightblue}{white}}}.                     \\ \hline
        5                 & Deadline decreases from 700 ms to 500 ms in {\small{\fullsmallcircled{4}{midnightblue}{midnightblue}{white}}}.  \\ \hline
        6                 & Deadline decreases from 500 ms to 250 ms in {\small{\fullsmallcircled{5}{midnightblue}{midnightblue}{white}}}.   \\ \hline
        7                 & VGG16-based app starts in {\small{\fullsmallcircled{4}{midnightblue}{midnightblue}{white}}}.                         \\ \hline
        8                 & Deadline decreases from 700 ms to 500 ms in {\small{\fullsmallcircled{7}{midnightblue}{midnightblue}{white}}}.  \\ \hline
        9                 & Deadline decreases from 500 ms to 250 ms in {\small{\fullsmallcircled{8}{midnightblue}{midnightblue}{white}}}.   \\ \hlineB{2}
        \end{tabular}
        \vspace*{-6mm} 
        \label{tbl:scenarios}
    \end{table}
\endgroup

We assume the initial environment 
({\small{\fullcircled{0}{midnightblue}{midnightblue}{white}}}) 
to be in a state where YOLO-v3 operates alone and there is 
enough time and memory space to run the full-precision model.
In such a case, \name\ assigns 1026 MiB to the app, 
expecting the highest accuracy when the 
accuracy profile is used. 
Although the target deadline (700 ms) is determined while considering 
the full-precision model, \name\ selects a different 
model (INT layer accounts for 21\% of the total)
which runs more quickly and accurately with less resources.
In {\small{\fullcircled{1}{midnightblue}{midnightblue}{white}}}, 
despite sharing resources with VGG16, YOLO-v3 uses the 
same bit configuration used in 
{\small{\fullcircled{0}{midnightblue}{midnightblue}{white}}}.
This phenomenon is attributed to 
the low memory usage and time complexity of VGG16.
Also, given the accuracy of VGG16 is almost constant 
regardless of the degree of compression (see Fig. ~\ref{fig:profiles}),
\name~compresses VGG16 maximally and allocates 
most resources to YOLO-v3.
However, in {\small{\fullcircled{4}{midnightblue}{midnightblue}{white}}}, 
the execution of ResNet50 requires compression of
both ResNet50 and YOLO-v3.
Specifically, the memory space allocated to YOLO-v3 decreases 
from 1017 MiB to 952 MiB, and hence YOLO-v3 uses more low 
precision layers (60\% of the total) than 
{\small{\fullcircled{1}{midnightblue}{midnightblue}{white}}}.
In {\small{\fullcircled{7}{midnightblue}{midnightblue}{white}}},
the degrees of compression of ResNet50 and YOLO-v3 are the same as in
{\small{\fullcircled{4}{midnightblue}{midnightblue}{white}}}
due to the low computation and memory costs of VGG16.

{\small{\fullcircled{2}{midnightblue}{midnightblue}{white}}}, 
{\small{\fullcircled{5}{midnightblue}{midnightblue}{white}}}, 
and {\small{\fullcircled{8}{midnightblue}{midnightblue}{white}}} 
show \name's resource optimization for concurrent apps when 
the deadlines are tightened (from 700 ms to 500 ms).
{\small{\fullcircled{3}{midnightblue}{midnightblue}{white}}}, 
{\small{\fullcircled{6}{midnightblue}{midnightblue}{white}}}, 
and {\small{\fullcircled{9}{midnightblue}{midnightblue}{white}}} 
show the optimization results with the deadlines tightened 
further (from 500 ms to 250 ms).
In these scenarios, concurrent DNN models are executed 
by building more compact mixed-precision models.
Consequently, a larger margin between the peak memory 
usage and memory capacity is made available in 
Fig.~\ref{fig:multi}~(a).
The overall margins between the measured latency/peak memory usage 
and deadline/memory capacity may seem large, 
because we used the worst-case latency and maximum peak memory 
usage (through iterative measurements) when profiling, 
and hence the margins vary with the execution.

\looseness=-1
Our resource allocation aims to maximize total accuracy
without violating any timing constraint,
and hence there is no significant loss in accuracy 
(see Fig.~\ref{fig:multi}~(c)).
Due to the lack of space, we presented the results of YOLO-v3 only, 
the most influential model for both real-time execution and total 
accuracy drop.
In all the scenarios considered, all three models show 
less than 2.2\% accuracy drops. 
The maximum accuracy drop occurs in YOLO-v3 in 
{\small{\fullcircled{9}{midnightblue}{midnightblue}{white}}}.

\begin{figure}[pt]
    \begin{center}
     \includegraphics[width=0.9\linewidth]{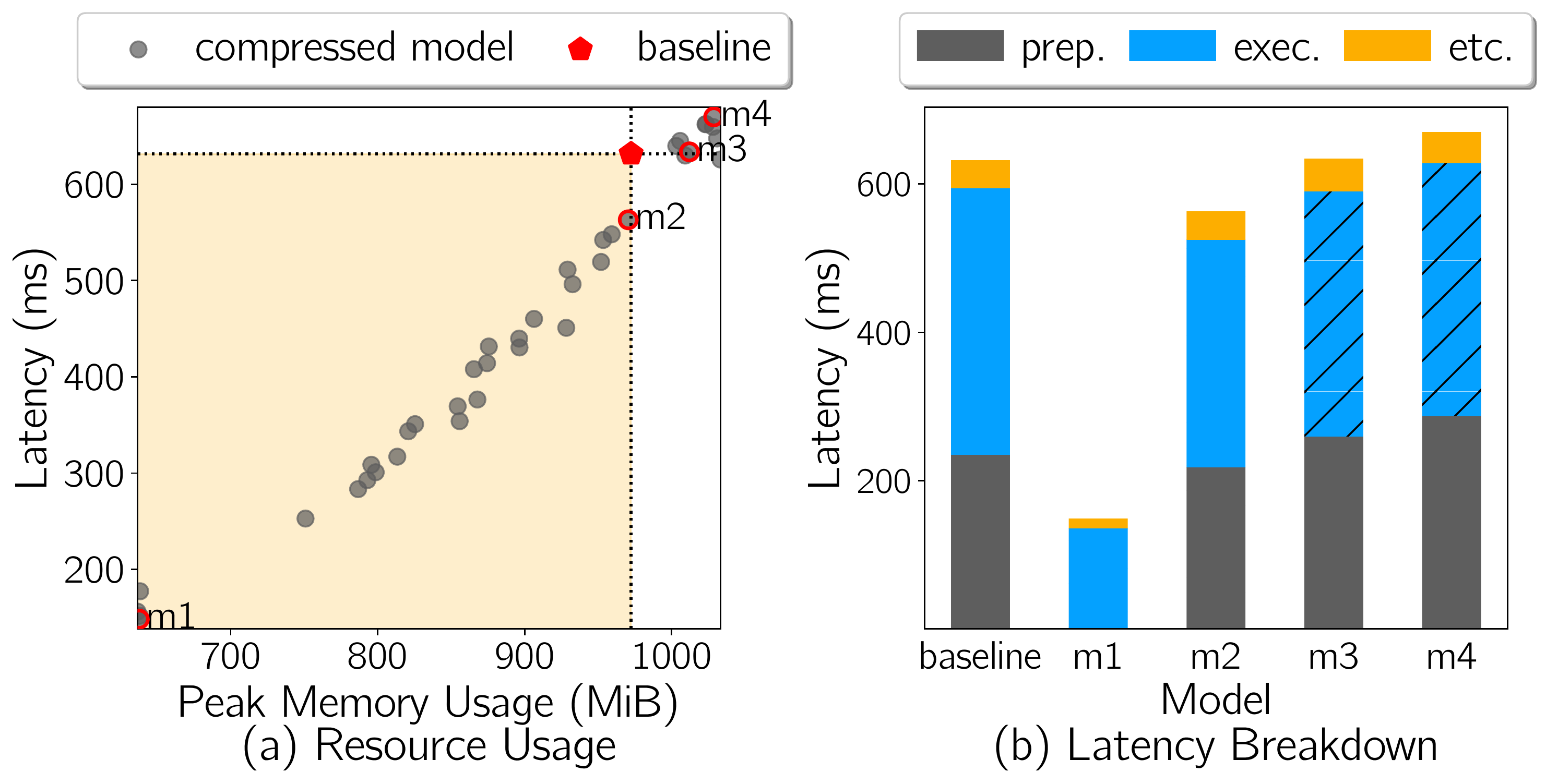} 
    \end{center}
    \vspace*{-4mm} 
    \caption{Comparison of results when \name~and the baseline are applied to YOLO-v3. In (a), the red pentagon represents the worst-case latency and peak memory usage when using the baseline, each of which is calculated using 100 measurements. (b) shows the latency breakdown of the main models in (a). 
    In the case of the baseline, ``prep.'' means the time required in loading the entire model;
    otherwise, it means the time for model reconfiguration.
    ``etc." includes the time spent in model restoration, interprocess communication.}
    \label{fig:baseline}
    \vspace*{-4mm} 
\end{figure}

\subsection{Performance Comparison}
\noindent\subsubsection{Comparison with baseline}
If a new app is allotted enough resources to run the full-precision 
model in \textbf{\em case~1}, the subprocess will reconfigure the 
DNN pipeline by loading all its FP layers, which may use resources 
less efficiently than the baseline approach (i.e., loading 
the entire FP model at once {\em without} reconfiguration). 
Albeit such inefficiency, we show \name~to make 
efficient use of resources in most cases.

Fig.~\ref{fig:baseline}~(a) shows the peak memory usage and 
worst-case latency of the mixed-precision models of YOLO-v3
(the same as the YOLO-v3 latency graph in 
Fig.~\ref{fig:profiles}), but this time, the peak memory usage 
and worst-case latency resulting from the baseline approach are 
marked with red pentagon, and the latency breakdowns for the 
main cases (marked with red hollow dots) are shown in Fig.~\ref{fig:baseline}~(b).
This analysis suggests several aspects of \name's effectiveness.
First, \textit{m4} (all layers in the execution path 
are replaced with FP layers) shows inefficient latency and 
memory usage compared to the baseline.
However, 73.7\% of our compressed models showed less 
usage of memory and processing time (the yellow area in Fig.~\ref{fig:baseline}), especially  78.9\% of 
our compressed models showing reduced latencies.
In case of \textit{m1} (all layers are in low precision), latency 
and memory usage were reduced by 4.3$\times$ and 1.5$\times$, respectively, compared to the baseline.
Additionally, when about 22.7\% of the layers were quantized 
(\textit{m3}), it showed the same level of latency as the 
baseline, and when about 30.7\% of the layers were quantized 
(\textit{m2}), it showed the same level of memory usage.
From these results, we conclude \name~reacts well to the given 
deadlines by optimizing the computation and memory resources required in running apps. 

\noindent\subsubsection{Comparison with NestDNN}
NestDNN \cite{fang2018nestdnn} also uses a DNN pruning method 
to run concurrent vision apps (non-real-time). 
They create a fixed number of small models according to 
the degree of pruning. They demonstrated feasibility in terms of 
memory usage, accuracy, and energy consumption.
Energy consumption is not in our scope and the app's memory usage 
is discussed by using its model size as an indirect metric. 
So, we compare \name\ with NestDNN only for accuracy.

\looseness=-1
Table~\ref{tbl:nestdnn} shows the differences between the 
generated models when the two methods are applied to VGG16.
Here, \name~is demonstrated: 1) for all the 
generated mixed-precision models (ALL) and 
2) for the models saved in the lookup table (LUT).
While NestDNN uses only five compressed models 
for online resource allocation, 
we can handle more models by inserting the app profiles
into the optimization problem and 
reducing the time of searching for models via 
lookup tables.
Furthermore, all of the compressed models generated by \name~show 
higher accuracy than NestDNN, because its pruning scheme
can incur significant accuracy drop since the distribution of 
model parameters is changed. 
Although the authors recover the 
accuracy to some extent via retraining, the accuracy is not 
fully recovered due to the diminished capacity of the model.
The higher accuracy of \name~enhances the feasibility, but 
shows a lower compression ratio.
However, the degree of compression can be increased by using 
lower precision (e.g., INT4), and if different low precisions 
are used simultaneously, we can increase both the compression 
degree and accuracy.

\begingroup
\setlength{\tabcolsep}{3pt}
\renewcommand{\arraystretch}{1.0}
\begin{table}[pt]
\centering
\caption {Comparison between NestDNN \cite{fang2018nestdnn} and \name. Max compression ratio means the ratio of the uncompressed model size to the smallest compressed model size.}
\vspace*{-4mm} 
\begin{tabular}{cccc}\\ \hlineB{2}
\textbf{}                                   &\cite{fang2018nestdnn}     & \name~(All)         & \name~(LUT)         \\ \hlineB{2}
\# Compressed models                        & 5                         & 17                   & 2                  \\ \cline{1-4}
Min accuracy                                & 83                        & 92.47               & 92.47               \\\cline{1-4}
Max accuracy                                & 89                        & 92.51               & 92.51               \\ \cline{1-4}
Max compression ratio                       & 7.8                       & 4.0                 & 4.0                 \\ \hlineB{2} 
\end{tabular}
\vspace*{-5mm} 
\label {tbl:nestdnn}
\end{table}
\endgroup

\subsection{Robustness of \name}
\label{sec:robustness}
Since system error in AVs can cause exigent situations, we must 
ensure system robustness in terms of system availability and 
quality --- two critical factors of system failure.
The resource optimization mechanism in \name\ fails to yield 
results if the deadline is set below a certain level.
Furthermore, heavily compressed models can be used to meet 
the short deadline but may lead to fatal accidents
due to their low accuracy. 
Considering these factors that affect 
system availability and quality, we can evaluate system robustness.

Below we show the robustness of \name\ using the total accuracy 
resulting from lookup tables.
Since the robustness may be different in the platforms with 
different memory capacity, we visualized the relationship between 
deadline, memory capacity, and total accuracy in 
Fig.~\ref{fig:robustness}.
The black cells indicate the points where system gives invalid 
results for three reasons: when available bit-widths cannot meet the 
deadline, when memory capacity cannot support \name, or when the 
total accuracy is degraded by more than a predefined threshold 
(3\% in this paper). In the remaining cases, \name\ guarantees 
acceptable quality for users.

As our environment supports INT8 and FP32 operations, total 
accuracy does not decrease by more than 3\%, but the 
lower bounds of deadline and memory capacity appear high
when the black area is considered.
The ultra-low-bit integers, such as INT4 and INT2, reduce the lower 
bounds, thus allowing \name\ to operate without errors under 
tighter deadlines and less memory capacity.
Therefore, we can improve system robustness by using the various 
bit-widths the system supports.

\begin{figure}[pt]
    \begin{center}
     \includegraphics[width=0.6\linewidth]{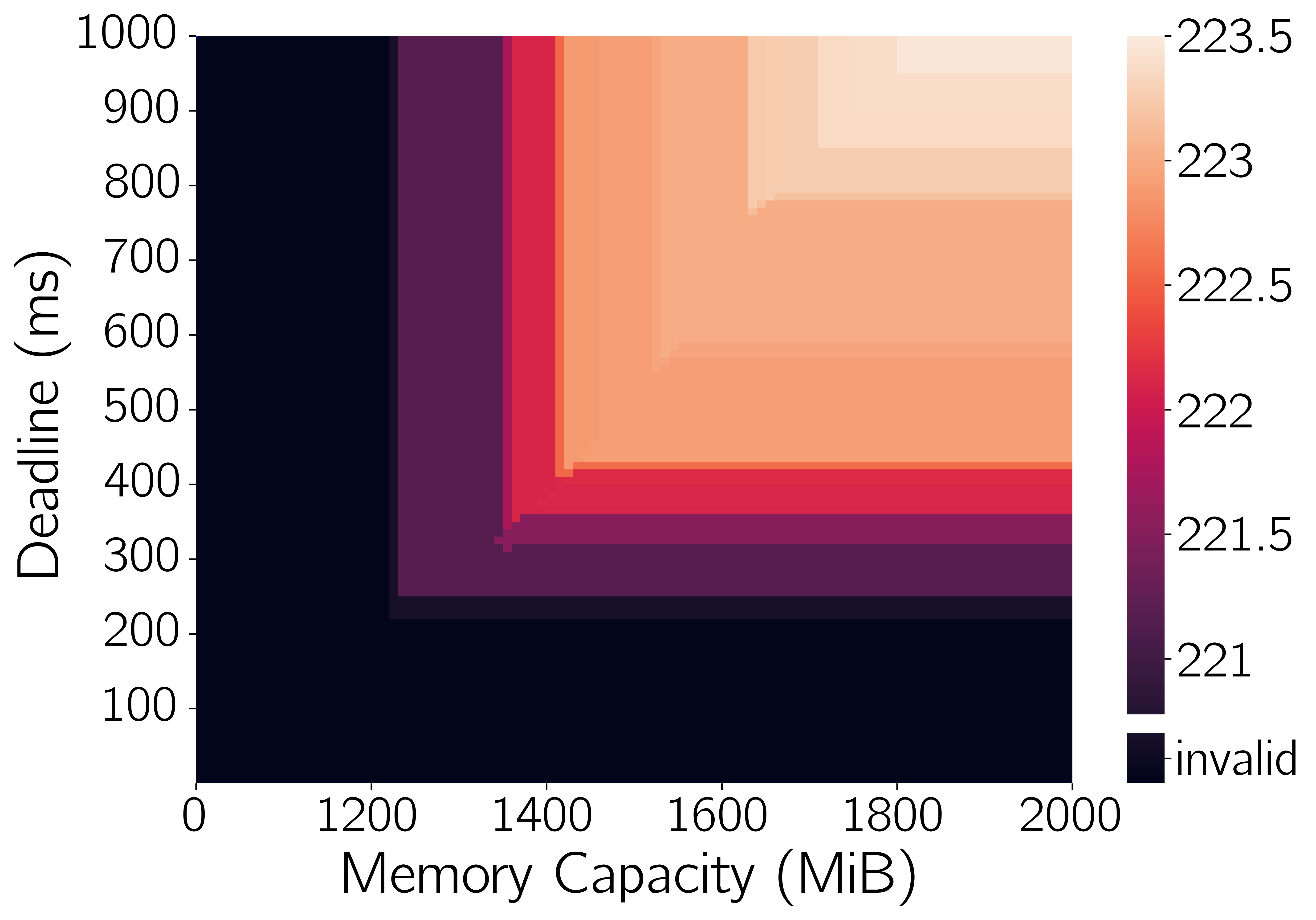} 
    \end{center}
    \vspace*{-5mm} 
    \caption{Total accuracy while changing the deadline and memory 
    capacity. The lighter the color, the higher the total accuracy. 
    Black cells denote invalid results caused by the inability of 
    optimization or a significant drop in accuracy.}
    \label{fig:robustness}
    \vspace*{-4mm} 
\end{figure}

\section{Conclusion and Future Work}
\label{sec:conclusion}
We have proposed a resource optimization framework, called \name,
that allows multiple DNN-based real-time apps to meet the same 
deadline of apps (dictated by camera's frame rate),
which is critically important for CPS apps.
\name~creates the app performance profiles offline,
and reconfigures the mixed-precision DNN models at runtime 
while reflecting dynamically varying resource requirements 
of the running apps according to their status so as to
share the platform resources.
Our implementation in a simulated environment has 
demonstrated its feasibility and effectiveness.

We assume enough computation resources
for the apps so that their models can fit in memory. 
How to relax this assumption is part of our future work.
We would also like to apply \name\
in mixed criticality system where multiple apps of different 
criticality levels can run.
Although its current version focuses on CNN-based apps, 
\name~can be extended to support various DNN-based apps including 
language models, facilitating the execution of many 
real-time apps on AVs.

\bibliographystyle{ieeetran}
\bibliography{ref}
\end{document}